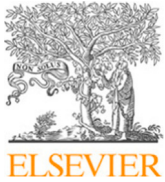
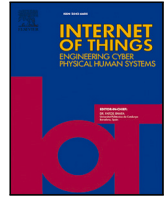

Contents lists available at ScienceDirect

# Internet of Things

journal homepage: www.elsevier.com/locate/iot

Research article

# Smartphone-based eye tracking system using edge intelligence and model optimisation

Nishan Gunawardena [*], Gough Yumu Lui, Jeewani Anupama Ginige, Bahman Javadi

*Western Sydney University, Locked Bag 1797, Penrith, 2751, NSW, Australia*

ABSTRACT

A significant limitation of current smartphone-based eye-tracking algorithms is their low accuracy when applied to video-type visual stimuli, as they are typically trained on static images. Also, the increasing demand for real-time interactive applications like games, VR, and AR on smartphones requires overcoming the limitations posed by resource constraints such as limited computational power, battery life, and network bandwidth. Therefore, we developed two new smartphone eye-tracking techniques for video-type visuals by combining Convolutional Neural Networks (CNN) with two different Recurrent Neural Networks (RNN), namely Long Short Term Memory (LSTM) and Gated Recurrent Unit (GRU). Our CNN+LSTM and CNN+GRU models achieved an average Root Mean Square Error of 0.955 cm and 1.091 cm, respectively. To address the computational constraints of smartphones, we developed an edge intelligence architecture to enhance the performance of smartphone-based eye tracking. We applied various optimisation methods like quantisation and pruning to deep learning models for better energy, CPU, and memory usage on edge devices, focusing on real-time processing. Using model quantisation, the model inference time in the CNN+LSTM and CNN+GRU models was reduced by 21.72% and 19.50%, respectively, on edge devices.

## 1. Introduction

Eye tracking is the process of measuring eye movements. It helps to determine where users are looking and for how long they fixate their gaze on a particular location. According to Global Market Insights, the Eye Tracking Market size was valued at US$ 852.6 million in 2023 [1], and it is expected to grow at a compound annual growth rate of over 30.5% between 2024 and 2032. This implies a significant rise in the integration of eye-tracking technology across various industries. The increasing demand for eye-tracking systems in sectors such as automotive, healthcare, retail, and consumer electronics drives this growth [1]. Additionally, the advancement in eye-tracking technology, such as improved accuracy, ease of integration, and affordability, further accelerates market expansion. As the market continues to grow, it opens up new opportunities for innovative applications in human–computer interaction, psychological research, and accessibility solutions, thereby shaping the future of eye-tracking technology.

Given the importance of eye tracking, it is necessary to examine how different types of visual stimuli on smartphones can affect human gaze estimation. Goldberg et al. [2] used functional magnetic resonance imaging to study attention-related brain activity when viewing dynamic and static visual stimuli. The study found that dynamic stimuli activated brain regions linked to attentional control more than static stimuli. Smartphone eye-tracking applications relying on user interaction with dynamic visual content, such as virtual reality (VR), augmented reality (AR), and gaming, require accurate and efficient eye-tracking algorithms to monitor gaze and ensure a positive user experience. These applications demand real-time processing and adaptability to various visual stimuli.

However, a significant limitation of current smartphone-based eye-tracking algorithms [3,4] is their reduced accuracy with video-type visual stimuli, as these algorithms are trained on static images. To address this challenge, we proposed two appearance-based eye-tracking models for smartphones that utilise the front-facing camera. Our approach combines CNN with Recurrent Neural Networks (RNN), specifically LSTM and GRU architectures.

Running a deep learning model for extended periods on a smartphone presents challenges due to resource constraints such as limited computational power, battery life, and network bandwidth, which affect real-time processing. To address these issues and improve the feasibility of the proposed eye-tracking algorithm on smartphones, we explore the use of edge intelligence. This approach shifts deep learning model inference from the device to nearby edge servers. To assess the performance of these algorithms in terms of processing efficiency and resource utilisation, we evaluated inference time, CPU and memory usage, and energy consumption on various edge devices, including Raspberry Pi 4 (RPi 4), Odroid N2+, Intel NUC, and Nvidia Jetson AGX.

The other objective of this study is to determine the level of performance improvement obtainable via the use of optimisation techniques such as pruning and quantisation and the magnitude of the accuracy trade-off that occurs. Pruning is the process of identifying and removing unnecessary weights from a neural network. In contrast, quantisation focuses on reducing the precision of the weights and activations in a neural network. Therefore, the proposed models were subjected to both pruning and quantisation to evaluate the effects of these techniques on their performance. By implementing these optimisations, we aimed to enhance the model's efficiency, making it more suitable for deployment on resource-constrained edge devices.

This paper makes the following contributions:

- Design and development of deep learning-based models for gaze estimation on smartphones for video content with high accuracy. (Section 3)
- Propose an edge intelligence architecture for smartphone-based eye tracking and evaluate the inference time, CPU and memory usage, and energy consumption with real edge devices. (Sections 4 and 5)
- Evaluation of two model optimisation techniques called quantisation and pruning to enhance the efficiency of the proposed eye-tracking models. (Sections 4 and 5)

The subsequent sections of this paper are organised as follows: Section 2 provides an overview of related methodologies and approaches. Section 3 includes the proposed deep learning architecture, the data collection and preprocessing methods, an implementation of the proposed deep learning architectures, and performance evaluations of the proposed models. Section 4 explores proposed edge intelligence architecture and deep learning model optimisation techniques such as pruning and quantisation. The results and findings of the experiments are discussed in Section 5. Finally, we draw a conclusion in Section 6.

## 2. Related work

This section provides an overview of previous research in eye tracking on smartphones and the integration of edge intelligence and model optimisation for real-time applications. It highlights the evolution of methodologies, challenges faced, and gaps that this study aims to address.

### 2.1. Eye tracking on smartphones

Traditional eye-tracking methods have faced challenges such as high costs, lack of portability, limited accuracy in dynamic environments, and extensive calibration needs [5–7]. The integration of advanced camera technologies and sophisticated machine learning algorithms in smartphones offers a promising avenue to address these challenges [3,4,8]. Over the years, researchers have explored various gaze estimation methodologies, predominantly focusing on model-based, appearance-based, and feature-based techniques. In model-based gaze estimation, two primary methods have been identified: shape-based and corneal-reflection techniques. Shape-based methods estimate gaze direction by analysing facial features, such as the shape of the eyeball, as demonstrated by Song et al. [9], or the geometric configuration of the pupil and eye corners, as developed by Ishikawa et al. [10]. Conversely, corneal-reflection methods utilise reflections within the eye, pinpointing gaze direction more accurately with the help of an external infrared light source, a technique noted by Duchowski [11] and widely used in commercial eye trackers for its precision.

The objective of appearance-based gaze estimation methods is to establish a direct mapping from images to gaze information. The enhancement of smartphone camera capabilities has led to increased research into appearance-based gaze estimation on mobile devices. Lei et al. [12] report that commonly used eye-tracking algorithms for smartphones rely on CNNs, which are generally employed for image classification tasks. Krafka et al. [3] applied an AlexNet-type CNN using face, face-grid, and two eyes as inputs to predict gaze location on a fixed red dot, achieving an average Root Mean Squared Error (RMSE) of 1.71 cm on mobile phones and 2.53 cm on tablets. Bâce et al. [13] developed two CNNs for face detection and gaze estimation, resulting in a Matthews Correlation Coefficient (MCC) [14] of 0.74. Valliappan et al. [4] employed a modified CNN architecture, adjusted with calibration data and supplemented by a regression model, resulting in an average RMSE of $1.92 \pm 0.20$ cm with 26 participants. These studies utilised the GazeCapture dataset [3], and the focus of these studies was mainly on static stimuli.

Human attention may change with the type of visual stimuli that are presented on the smartphone. For instance, dynamic visual stimuli, like videos and mobile games, require more attention than static visual stimuli, such as web pages and images [15]. The primary distinction between static and dynamic stimuli is the inclusion of motion and temporal variations in the latter, which introduce additional complexities in gaze estimation. However, Mark et al. [16] used a pre-recorded 30-s video to assess the perception of high-stress scenarios in an aviation mission simulator. In [17,18], scholars have used eye tracking with video stimuli to assess participants' attention in pre-learning unknown words and visual language. These studies indicate that eye tracking can provide useful insights into understanding humans' gaze behaviour when looking at unfamiliar video-type dynamic visuals. All these studies were performed using commercial eye trackers, and we have identified only one study that used both appearance-based mobile device eye tracking and a commercial eye tracker with a video stimulus. Ottoom et al. [19] developed an appearance-based gaze estimation system for smartphones, aiming to determine which of the nine predefined regions on the screen a user is looking at while watching an American Sign Language video. They have achieved 91.3% accuracy compared to the commercial eye tracker. Therefore, we see a clear research gap in appearance-based eye-tracking algorithms in smartphones for dynamic visual stimuli enabling real-time gaze estimation on videos, games, or interactive content on smartphones. While previous studies focused on static stimuli, our approach specifically addresses the challenges posed by dynamic visual content on smartphone-based eye tracking.

There is limited research exploring the combination of CNN and RNN architectures for developing appearance-based eye-tracking systems. The advantage of employing both CNN and RNN lies in the ability to harness their respective strengths, ultimately leading to enhanced accuracy in eye tracking. CNNs excel at learning spatial features within images, while RNNs are proficient in capturing temporal dependencies in sequential data. Palmero et al. [20] introduced a gaze estimation algorithm based on CNN+LSTM for a 24″ screen, resulting in a 4% accuracy improvement compared to a static CNN approach. Park et al. [21] presented a gaze estimation algorithm utilising a CNN+GRU model, achieving a 28% performance enhancement in gaze direction estimation error compared to alternative models. However, none of these studies were performed in the context of eye tracking on a handheld mobile device.

## 2.2. Edge intelligence and model optimisation for eye tracking

Smartphones have witnessed a surge in processing capabilities in recent years; however, they remain constrained by battery life and computational resources. This poses a challenge in executing real-time and compute-intensive applications. In such cases, offloading these tasks to other devices, such as edge devices or cloud computing platforms, becomes necessary [22]. Particularly, transferring the computationally intensive eye-tracking applications from smartphones to either cloud or edge infrastructure can substantially reduce computational power, memory usage, and power consumption on the smartphone. This, in turn, contributes to enhancing the overall quality of experience.

Dao [23] explored using cloud computing systems for real-time eye-tracking analysis for fixation detection, heat map visualisation, and eye-tracking data classification. Dao's findings demonstrated that cloud computing offers advantages over single PCs in terms of running time, data aggregation, and scalability. However, cloud computing has some limitations in terms of latency, network bandwidth, and privacy.

The need for edge intelligence [24] becomes apparent in light of these limitations. In edge intelligence, artificial intelligence (AI) algorithms, such as machine learning models, are deployed on edge devices to perform tasks such as data analysis, pattern recognition, and decision-making in real time. This brings computation and AI-driven data processing closer to the data source, reducing latency and enabling rapid, real-time decision-making. Therefore, edge intelligence provides immediate decision-making capabilities and enhances data privacy and security by minimising data transfer to centralised servers. Moreover, it offloads the cloud infrastructure, reducing internet bandwidth and transit costs.

Future applications of eye tracking technology may require integration with more real-time applications such as augmented reality (AR) systems and mobile games [25]. However, these integrations demand high-performance computational platforms to handle eye-tracking data efficiently and provide feedback in real-time [26]. In our previous study [27], we compared four lightweight CNN models for smartphone-based eye tracking using different inference modes, including on-device, cloud-based, and edge-based. The results of this study proved that on-device inference is constrained by energy consumption and memory usage, and the high communication time between smartphones and the cloud can make real-time eye-tracking applications impractical with cloud-based inference. As a result, edge intelligence becomes a more effective solution for delivering a higher-quality user experience in smartphone-based eye-tracking applications.

Privacy is another important factor in eye-tracking applications as they capture the user's facial features. Edge intelligence can help to reduce the risk of privacy breaches by keeping data local and reducing the need to transmit it to the cloud. This can be done by processing the eye-tracking data on the device itself or on a nearby edge device. Edge devices can also encrypt the data before it is sent to the cloud or stored in a secure location. Considering the advantages of edge intelligence, including low latency, increased security and privacy, low cost, and improved bandwidth usage, we propose an edge intelligence-based system architecture for smartphone-based eye-tracking applications.

Model optimisation is crucial for deploying deep learning models on edge devices, focusing on techniques like quantisation, pruning, and compression to reduce model size and resource requirements. Chen et al. [28] achieved a 4.25% accuracy improvement and 3.5x to 6.4x memory reduction in biomedical image segmentation using deep learning model quantisation. Additionally, Han et al. [29] reduced the VGG-16 model size by 49x with no loss of accuracy using pruning, quantisation, and Huffman coding. These results underscore the significant impact of model optimisation in enhancing performance and efficiency for edge intelligence applications, especially in resource-constrained environments.

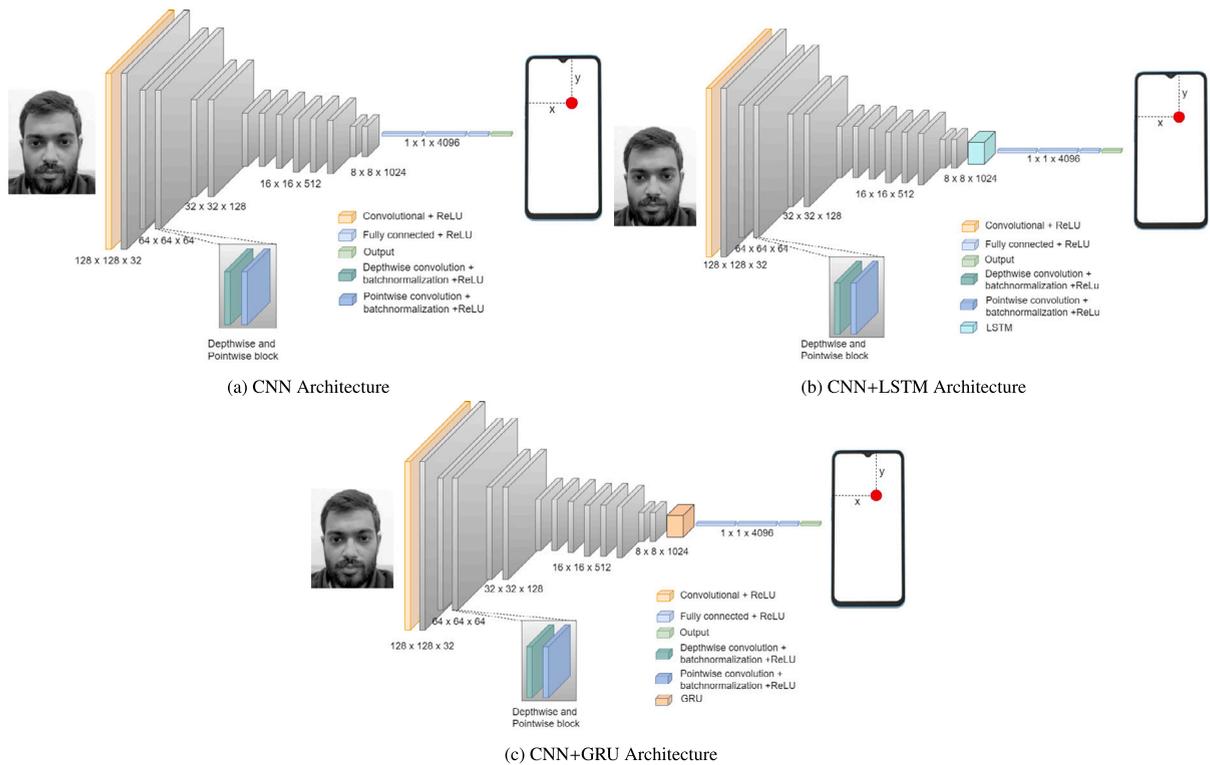

**Fig. 1.** Architectural design of the proposed three deep learning models. The input to the models is greyscale images of the user's face captured using the front-facing camera of a smartphone. The output of the models is a pair of continuous values (x, y) that represent the predicted coordinates of the user's gaze point on the smartphone screen.

## 3. Deep learning-based eye tracking for smartphones

This section presents the deep learning architectures used in this study for eye tracking on smartphones with dynamic stimuli. It includes the design and implementation details of three different models, compares their performance, and discusses the results in terms of accuracy and on-device inference time. The section also introduces the data collection methods used to create a new dataset, which addresses the lack of dynamic visual stimuli in existing datasets.

In this study, four deep learning architectures were used for eye tracking on smartphones with dynamic stimuli. The initial architecture, the iTracker [3], served as the baseline for comparison. The other three architectures combined CNN and RNN layers. Fig. 1 presents the architectures used in this study. The iTracker model is derived from the AlexNet architecture [30] and adapted for gaze estimation. It incorporates inputs from a face, two eyes, and a face grid. Trained using the GazeCapture dataset, the iTracker model registers an error of 1.71 cm on iPhones without requiring calibration. A streamlined version of this model operates in approximately 50 ms on an iPhone 6s. When integrated with Apple's face detection technology, it achieves a detection rate of 10–15 frames per second on an iPhone. This architecture serves as the benchmark for comparison in this research.

### 3.0.1. CNN

In our prior study [27], we analysed several CNN architectures such as AlexNet, LeNet-5, ShuffleNet-V2, and MobileNet-V3 for eye tracking on smartphones using static stimuli. The results indicated that the MobileNet-V3 architecture surpassed the others in accuracy and computational efficiency. Specifically, MobileNet-V3 recorded the lowest RMSE at 1.42 cm and demonstrated an inference time of only 65 ms per frame when evaluated using the GazeCapture dataset. The first proposed architecture we used in this study is based on the findings of our previous study [27].

We used the concept of depthwise and pointwise convolutions from the MobileNet-V3 architecture to enhance efficiency, reduce complexity, and increase the non-linearity of the models proposed for dynamic stimuli. Depthwise convolution involves applying a separate filter to each input channel, allowing the filter to learn distinct features from each channel, which can enhance performance. Pointwise convolution applies a single filter across all input channels, enabling the filter to learn features common to all channels, thereby also improving performance. By combining depthwise and pointwise convolutions, depthwise separable convolutions are formed. This technique first applies a depthwise convolution, followed by a pointwise convolution, improving performance while reducing the number of parameters. The architecture used in this study is presented in Fig. 1(a).

Table 1
Characteristics of the DynamicGaze dataset used in this study.

| Feature | DynamicGaze | GazeCapture |
| --- | --- | --- |
| Participants | 173 | 1474 |
| Device models | 42 | n/a |
| Number of frames | 67,591 | 2,445,504 |
| Min. screen Resolution | 720 × 1280 | n/a |
| Max. screen Resolution | 1440 × 3200 | n/a |
| Stimuli | dynamic (random) | static dots |
| Reference point | top-left corner of the screen | front camera |
| Platforms | iOS/Android | iOS |

### 3.0.2. CNN+LSTM

Combining CNN and RNN architectures enables the model to learn both spatial and temporal features of eye-tracking data. Long Short-Term Memory (LSTM) architecture, a type of RNN, is designed to mitigate the vanishing gradient problem often encountered in traditional RNNs. This issue arises when the gradient of the loss function becomes too small during backpropagation, hindering the model's ability to learn long-term dependencies in sequential data. LSTM differs from Gated Recurrent Unit (GRU) in its approach to managing past input memory. LSTM utilises three gates: input, output, and forget, to regulate information flow into and out of the cell state, allowing selective retention or deletion of prior input information. In contrast, GRU uses two gates: update and reset, to manage memory retention and updates, making it more computationally efficient than LSTM. Previous studies have demonstrated the effectiveness of the CNN+LSTM model. For example, Srinivasu et al. [31] achieved 85.34% accuracy using the CNN+LSTM architecture for skin disease classification and detection with minimal computational power.

In this study, the CNN architecture described in Section 3.0.1 is used to extract features from the input, while the LSTM captures and models the temporal dependencies between the CNN's output vector and the corresponding gaze coordinates over time. This approach improves the model's ability to capture the relationship between eye movements and gaze points, leading to more accurate predictions. The LSTM layer processes the sequence of gaze points over time and outputs a fixed-length representation of the temporal context. By incorporating the LSTM layer, the model better captures the temporal dynamics of eye movements, enhancing prediction accuracy. To the authors' knowledge, this is the first study to combine the CNN+LSTM approach for smartphone-based eye tracking.

### 3.0.3. CNN+GRU

Compared to LSTM, GRU uses gating mechanisms that regulate information flow more efficiently, leading to improved performance in modelling short-term dependencies. Specifically, GRU employs two gates: reset and update gates. The reset gate controls how much of the previous state is forgotten, while the update gate determines the new information to incorporate into the current state. This simplified gating structure allows GRU to manage short-term dependencies more effectively than LSTM. For instance, in short-term residential load forecasting, Sajjad et al. [32] utilised a CNN+GRU-based approach, achieving the lowest mean squared error (MSE) of 0.09 compared to other models, including CNN+LSTM.

In this study, we also combined the CNN architecture described in Section 3.0.1 with a GRU layer as the fourth architecture. Although the CNN+LSTM model has shown promising results in prior studies, we aimed to assess the performance of a more lightweight model with fewer parameters to identify which approach is more suitable for eye tracking on devices with limited resources.

### 3.1. Data collection

We have made use of two different datasets. The initial dataset used is the publicly accessible GazeCapture dataset [3]. We have created another dataset named DynamicGaze dataset to overcome the unavailability of smartphone-based eye-tracking data with dynamic visual stimuli and the lack of diversity in smartphones in the GazeCapture dataset. Table 1 summarises the key characteristics of both datasets.

We developed a web-based application to collect data from various types of smartphones. We utilised a moving dot that traversed random paths across the entire screen to represent the dynamic nature of the stimuli. Most crowdsourcing platforms do not optimise for smartphones, leading us to develop a web-based application specifically for smartphones and tablets. For participant recruitment, we utilised the Amazon Mechanical Turk (AMT)[1] platform and provided participants with instructions on how to use the application. The application was hosted on the Western Sydney University servers[2] to ensure data security and research ethics compliance.

Participants were asked to contribute three videos, each taken with their smartphone in a different position relative to their eyes (below eye level, at eye level, and above eye level). The diverse range of device models and screen resolutions ensures that our models are more robust and can generalise better across different devices. Participants were compensated USD 0.50 for each set of three 20-s video recordings, which took approximately three minutes to complete. This study was conducted under the ethics approval of the Western Sydney University ethics committee (Ethics Approval Number — H14493). Once the participant has started the recording, they have to follow the circle with their eyes without moving their heads. After each video, they had to wait for the recorded video to upload before proceeding to the next one. The data was uploaded as videos, with metadata including user-agent strings, screen resolution, motion sensor data, and device pixel ratios. Upon completion, workers received a unique study code that they needed to submit for credit.

---

[1] https://www.mturk.com.
[2] https://mobileeyetracker.cdms.westernsydney.edu.au.

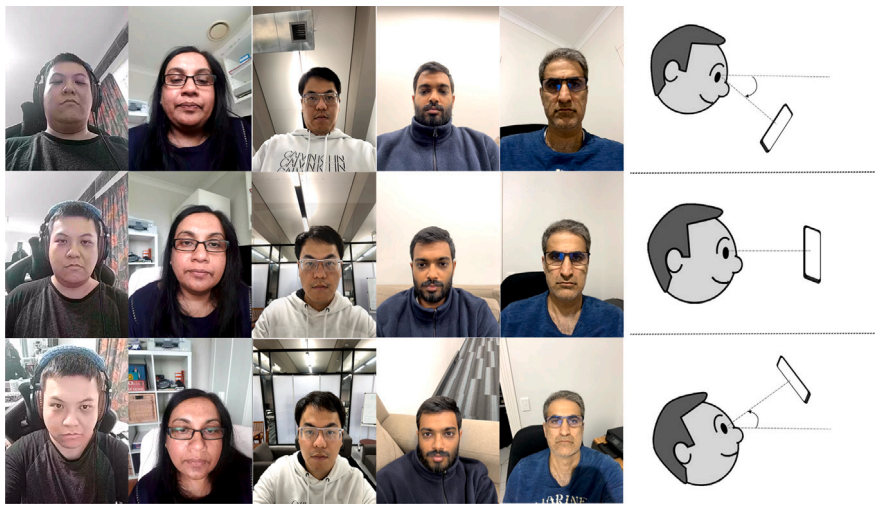

**Fig. 2.** Sample frames from the DynamicGaze dataset collected in three different positions including below eye level (first row), same as eye level (second row), and above eye level (third row). The individuals depicted in the figure have expressly consented to the publication of their faces.

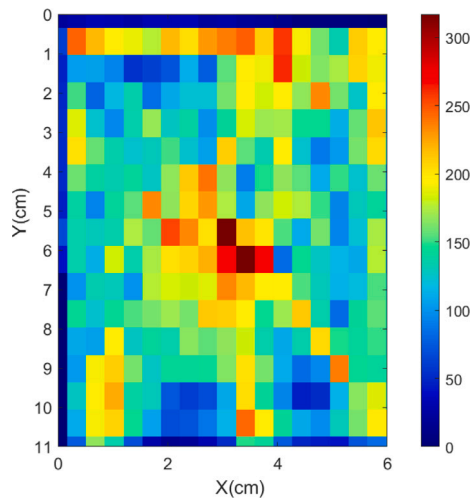

**Fig. 3.** Distribution of coordinates of the DynamicGaze dataset. Axes indicate centimetres from the top left corner of the screen. For example, the coordinates of the dot on the screen are projected to this space where the top left corner is at (0, 0).

Once the data was submitted, all the submitted data were examined carefully before acceptance. Fig. 2 visualises a set of frames from the DynamicGaze dataset, illustrating the differences in each position, appearance, background, and illumination. A moving dot following random paths covering the entire screen, introduces unpredictability and complexity, resembling real-world scenarios where gaze behaviour is more dynamic. This dataset challenges the algorithm's ability to adapt to changing stimuli and assesses its robustness and generalisability. Fig. 3 displays the distribution of coordinates in the DynamicGaze dataset with the top left corner of the screen located at (0,0).

### 3.2. Implementation

Following a standard image processing pipeline for gaze tracking, the video sequences were first converted into frames. Videos are essentially a sequence of frames displayed rapidly to create the illusion of motion. By converting videos into frames, the continuous stream of the video was broken down into individual frames, allowing us to analyse each frame independently. This breakdown enabled us to apply various image processing techniques, such as face detection and feature extraction, to each frame. Subsequently, face detection was performed using the Dlib HoG face detection algorithm, as it helps locate the face within each frame of the video, which is crucial for determining where the person is looking.

After face detection, the frames were converted to greyscale to simplify the data and reduce computational requirements. This simplification can lead to faster processing times and lower computational costs, making the algorithm more efficient, especially when processing large amounts of data. Colour information can be useful in certain contexts, such as detecting specific features or patterns. Gaze tracking primarily depends on the position and movement of the eyes [11], which can be effectively captured in greyscale.

All frames were then normalised and resized to a fixed size of 128 $x$ 128 pixels. This normalisation ensured that input features were on a similar scale, which helped the deep learning models converge faster during training and improved their overall performance. It also helps avoid numerical instability that can arise when working with data with a wide range of values. Resizing the frames to a fixed size, such as 128 $x$ 128 pixels, is beneficial for consistency and efficiency in processing. It standardises the input size for all frames, making designing and training the deep-learning models easier. Additionally, a fixed size reduces the computational complexity of the models, as they only need to process images of a consistent size, leading to faster inference times and more efficient use of resources. The choice of 128 $x$ 128 pixels for resizing is based on a balance between image quality and computational efficiency.

The coordinates of the moving circle, representing participants' gaze locations, were stored in the database as Cascading Style Sheets pixel values relative to the top left corner of the screen. Using the pixels per inch and device pixel ratio values, the physical distance for the coordinates in inches and centimetres relative to the top left corner (0,0) was calculated. The last step of the preprocessing is mapping coordinates into frames to ensure each frame in the video has a corresponding gaze position. One of the challenges encountered during this step was the variability in the number of coordinates and frames in the video within a recording. The main reason for this challenge is using different smartphones with varying frame rates. Each device may capture a different number of frames per second (fps), leading to variations in the total number of frames during a recording. The coordinates were generated at 20 coordinates per second, generating 400 coordinates per recording. Mapping between coordinates and interpolating corresponding frames was achieved using the presentation time stamp assigned to each frame and the timestamp of each coordinate.

Once the data is processed, the four deep learning models were implemented using the TensorFlow framework [33] and trained on the two datasets mentioned earlier. The training procedure included a batch size of 32 for 1,564,320 iterations on the GazeCapture dataset and 45,540 iterations on the DynamicGaze dataset, with a training/testing ratio of 80–20. The Adam optimiser with a learning rate of 0.001 and a decay rate of 0.0001 was used for all models. The training was conducted on an NVIDIA Grid RTX6000p-6Q GPU with 6 GB memory. The code was developed using Python 3.9 and executed on a Windows 10 machine with 32 GB RAM and an Intel Xeon Gold 5215 CPU.

Consistent with prior work [3], the model's performance is quantified in terms of the mean Euclidean distance (in centimetres) from the true fixation location, assumed to be the location of the moving dot on the screen at that instant in time. The efficacy of each model is assessed using two metrics: the RMSE in centimetres and the $R^2$ value. To ensure a robust and reliable evaluation of the models, k-fold cross-validation with $k = 5$ is employed. This approach allows us to efficiently utilise our DynamicGaze dataset by ensuring that each data point is used for both training and validation. Further, this will reduce the potential bias from a single train-test split and gain a more comprehensive assessment of the model's generalisation capabilities.

### 3.3. Results and discussion

#### 3.3.1. Accuracy

Table 2 presents a comparison of the four models (iTracker, CNN, CNN+GRU, and CNN+LSTM) evaluated on GazeCapture and the DynamicGaze dataset. These results suggest that the iTracker model with 6.780 million parameters performs differently depending on the dataset, with better performance on DynamicGaze compared to GazeCapture. The CNN model with 3.21 million parameters achieves RMSE values of 1.671 cm on GazeCapture and 1.468 cm on DynamicGaze. This comparison implies that a higher number of parameters does not necessarily result in better performance. In contrast, the CNN+LSTM and CNN+GRU architectures outperformed both CNN and iTracker architectures. Specifically, the CNN+LSTM model achieved the lowest RMSE of 0.955 cm on the DynamicGaze dataset, and the CNN+GRU model achieved the lowest RMSE of 1.091 cm on the DynamicGaze dataset.

Among the proposed models, the CNN model exhibits lower accuracy across all two datasets due to its fewer parameters than the other two models. However, the primary reason for selecting the CNN model with fewer parameters is to enable its deployment on resource-constrained mobile devices. The reduced parameter count significantly decreases training time, particularly when dealing with large datasets. The CNN+LSTM model has more parameters than the CNN+GRU model, as the GRU architecture incorporates fewer gates and parameters than LSTM.

The CNN+LSTM and CNN+GRU models outperformed the CNN model because they could better capture the temporal dynamics of eye movements. The LSTM and GRU components could track the eye's movement over time, allowing the model to make more accurate predictions about where the user was looking. In DynamicGaze, the CNN+LSTM model outperformed other models, including the CNN+GRU model. This performance difference could be due to the specific characteristics of DynamicGaze, which involved a moving dot following random paths covering the entire screen. This is because the LSTM's ability to retain information over longer sequences and manage long-term dependencies is more effective in tracking the moving dot's random paths. Furthermore, The LSTM's separate cell state and hidden state offer more control over memory retention, which enhances its performance in capturing complex patterns and dependencies in the data.

**Table 2**
Performance comparison of four deep learning models (iTracker, CNN, CNN+GRU, CNN+LSTM) on GazeCapture and DynamicGaze, including k-fold cross-validated RMSE (cm), $R^2$, and Inference Time (ms).

| Model | Dataset | Parameters | RMSE (cm) | $R^2$ | Inference (ms) |
|---|---|---|---|---|---|
| iTracker | GazeCapture | 6.780M | 1.821 ± 0.232 | 0.86 ± 0.045 | 742.22 ± 17.47 |
|  | DynamicGaze |  | 1.499 ± 0.210 | 0.83 ± 0.055 |  |
| CNN | GazeCapture | 3.210M | 1.671 ± 0.222 | 0.79 ± 0.050 | **255.67 ± 8.11** |
|  | DynamicGaze |  | 1.468 ± 0.195 | 0.87 ± 0.045 |  |
| CNN+GRU | GazeCapture | 3.343M | **1.632 ± 0.213** | 0.82 ± 0.050 | 414.81 ± 9.66 |
|  | DynamicGaze |  | 1.091 ± 0.130 | 0.80 ± 0.035 |  |
| CNN+LSTM | GazeCapture | 3.344M | 1.642 ± 0.206 | 0.85 ± 0.045 | 426.18 ± 10.24 |
|  | DynamicGaze |  | **0.955 ± 0.139** | 0.81 ± 0.040 |  |

We further conducted an ANOVA test followed by Tukey's HSD test to evaluate the performance of these four deep learning models on DynamicGaze. The ANOVA test revealed a significant difference in the RMSE values among the models ($F(3, 396)$ = 254.42, $p < 0.001$). The Tukey's HSD test further identified that CNN+LSTM (mean RMSE = 0.97) performed significantly better than CNN (mean RMSE = 1.47), CNN+GRU (mean RMSE = 1.10), and iTracker (mean RMSE = 1.48). These results indicate that CNN+LSTM is the most effective model for these two datasets, significantly outperforming the others.

*3.3.2. On-device inference time*

Inference time measurements are important to determine if a model's performance can meet the demands of real-time applications. Therefore, the inference time for each model, measured in milliseconds, was recorded to evaluate their real-time capabilities. This assessment was conducted on a Samsung S22 smartphone using a Flutter application integrated with TensorFlow Lite models. Each model underwent five separate evaluations to confirm the consistency and dependability of the results. Table 2 collates the average inference time for processing a single frame, along with the standard deviation for each model.

The data reveals a distinct correlation between the complexity of a model and its computational efficiency. The iTracker model, which has the highest number of parameters at 6.780 million, also had the longest average inference time at 742.22 ms, suggesting that its complexity reduces its performance in real-time scenarios. Also, the CNN model, with fewer parameters at 3.210 million, registered the shortest average inference time of 255.67 ms, indicating its higher suitability for real-time applications due to quicker processing times. The hybrid models, CNN+GRU and CNN+LSTM, with parameters around 3.343 million and 3.344 million, respectively, displayed intermediate inference times of 414.81 ms and 426.18 ms.

It is important to note that these inference time measurements only account for the model's computation time and do not include other processing steps such as frame capture, face detection, and image preprocessing. Nevertheless, the reported times suggest that achieving real-time eye tracking with these models on current mobile devices remains challenging. Moreover, additional results from this experiment can be found in our previous paper [34].

## 4. The edge intelligence architecture for eye tracking

This section introduces the edge intelligence architecture designed for real-time smartphone-based eye-tracking applications. The section also discusses the proposed model optimisation techniques, including quantisation and pruning, to enhance the efficiency of deep learning models on resource-constrained devices. The experimental setup for evaluating these models on different edge devices is presented.

*4.1. The proposed architecture*

Considering the advantages of edge intelligence, we propose an edge intelligence architecture for smartphone-based eye-tracking applications as depicted in Fig. 4. The proposed architecture is a three-layered approach that includes the presentation, edge, and cloud layers.

The cloud layer encompasses a scalable cloud server equipped with high-performance resources capable of training deep neural networks. Leveraging GPU processing power, this server efficiently handles large datasets for training algorithms, thus enabling the development of newer models as more data becomes available. It also features data storage for preserving training data and eye-tracking data, such as gaze heatmaps.

The edge layer retrieves the pre-trained deep neural network model from the cloud data storage. This layer encompasses a range of devices, from low-power Raspberry Pi to more powerful edge devices, contingent on the specific eye-tracking application's requirements. Once the edge layer receives the video stream from the presentation layer, it initiates preprocessing and inference operations to deliver real-time predictions of gaze coordinates promptly. The results are conveyed back to the presentation layer, while gaze data is periodically stored on the cloud server for domain experts to access.

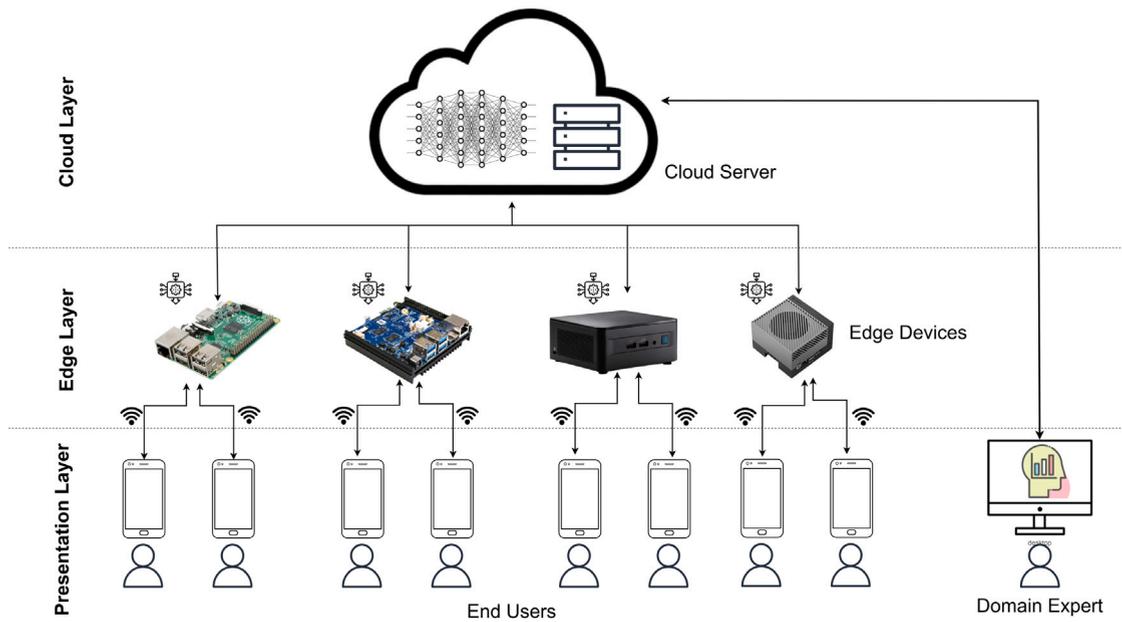

**Fig. 4.** Proposed edge intelligence architecture for real-time smartphone-based eye tracking applications.

The presentation layer is the bridge connecting end-users to various real-time eye-tracking applications, including augmented reality (AR) applications and healthcare and medical diagnosis tools. These applications capture a user's face and eye movements through the smartphone's front-facing camera. The captured video stream is then transmitted wirelessly to the edge layer for gaze direction prediction. Additionally, a cloud server is integrated to address requests from domain experts, who ensure the system's output aligns with practical requirements and domain-specific knowledge. They analyse gaze data visualisations (e.g., heatmaps) and interpret insights for applications in fields like healthcare diagnostics, user experience testing, or educational tools. Their feedback helps refine the system's accuracy and applicability by providing contextual understanding that may not be captured by automated processes alone.

### 4.2. Model optimisation

Optimising deep learning models in the context of edge intelligence is crucial to harness the full potential of this technology. Edge devices often have limited computational resources, including processing power and memory. Deep Learning models, known for their complexity and size, can strain these resources, leading to performance bottlenecks, increased energy consumption, and longer response times. Deep learning model optimisation addresses these challenges by tailoring models to operate efficiently on edge devices through model quantisation, pruning, and compression, which reduce model size without significantly sacrificing accuracy. The benefits of deep learning model optimisation in edge intelligence are manifold, including lower energy consumption, improved real-time processing, and the ability to deploy AI-driven applications on resource-constrained devices, ultimately enhancing the user experience. In this study, we employed model quantisation and pruning to enhance the efficiency of the three deep-learning models.

#### 4.2.1. Quantisation

Deep learning model quantisation reduces the precision of a model's weights and activations by representing them with fewer bits [35]. In deep learning models, weights and activations are typically stored as 32-bit floating-point numbers, which offer high precision but require more memory and computational resources. When quantisation is applied, these 32-bit floating-point numbers are converted to lower precision formats, such as 16-bit floating-point numbers or 8-bit integers. This reduction in precision allows the model to use less memory and perform computations faster, making it more efficient for deployment on resource-constrained devices like smartphones or edge devices.

The process of quantisation involves several steps. Quantisation can be done in two ways: post-training quantisation and quantisation-aware training. Post-training quantisation involves training the model in high precision and then converting it to low precision after training. During the quantisation-aware training, models will be trained using low-precision values, allowing the model to learn to compensate for the quantisation errors. During quantisation, the range of values that each weight or activation can take is mapped to smaller values using techniques such as linear or logarithmic scaling.

Quantisation helps to improve the inference time of deep learning models as lower precision data types require fewer bits for storage and processing. Additionally, quantised models use less memory, benefiting edge devices with limited memory resources. The lower precision arithmetic operations require fewer computational resources, and this can reduce CPU usage significantly. This will also decrease the energy required for computations.

**Table 3**
Specification of tested edge devices for evaluating developed eye tracking models.

| Specification | RPi 4 | Odroid N2+ | Intel NUC | Nvidia Jetson AGX |
| --- | --- | --- | --- | --- |
| CPU | Broadcom BCM2711 quad-core Cortex-A72 | Quad-core Cortex-A73 Dual-core Cortex-A53 | Intel Core i7-10710U | 8-core Arm® Cortex®-A78AE |
| Clock Rate | 1.5 GHz | 2.4 GHz | 1.10 GHz–4.70 GHz | 2.2 GHz |
| GPU | – | Mali-G52 (850 MHz) | Intel UHD Graphics 630 | NVIDIA Ampere |
| Memory | LPDDR4 8 GB | DDR4 4 GB | DDR4 16 GB | LPDDR5 32 GB |
| Storage | microSD | microSD | SSD | eMMC |
| OS | Ubuntu 22.04 | Ubuntu 22.04 (Mate) | Windows 11 | JetPack 5.1.2 (Linux) |

*4.2.2. Pruning*

Pruning is a process in which unimportant weights in a neural network model are identified and removed, particularly those with small values. This process typically involves iterative training of the model and then identifying and removing the least important weights based on magnitude or connectivity [36]. This iterative process involves cycles of pruning and fine-tuning to ensure that the pruned model retains its accuracy.

Removing these unimportant weights can significantly reduce the model's overall size, reducing memory requirements. This reduction in memory usage can also result in more efficient data transfer and storage. With fewer weights and neurons to process, the computational load decreases, leading to faster inference times. Moreover, pruning decreases CPU and energy consumption as it requires fewer computations than the baseline model.

Both quantisation and pruning aim to make models more efficient but in different ways. Quantisation provides immediate benefits in terms of reduced memory and faster inference due to simpler arithmetic operations. In contrast, pruning reduces the overall model complexity and can be more effective in lowering the number of computations. However, there is a trade-off between accuracy and efficiency. Quantisation can introduce quantisation errors due to the lower precision. Pruning can also affect the model's accuracy due to the removal of weights or neurons. Therefore, balancing accuracy and efficiency is crucial when applying these techniques.

To implement the quantised model, we converted the original TensorFlow model from a 32-bit floating-point representation, the baseline model, into a 16-bit floating-point TensorFlow Lite model using the post-training quantisation. During this process, the TensorFlow Lite Converter tool is utilised to transform the model, where the weights and activations are rounded or truncated to fit into the lower precision format. This transformation yielded a considerable reduction in the model size, resulting in a reduction of 74.2% for the CNN model, 69.3% for the CNN+GRU model, and 69.19% for the CNN+LSTM model.

Then, we converted the 32-bit TensorFlow model to a pruned model using a 0.1% pruning rate and the low magnitude method. The pruning schedule gradually increased the sparsity from an initial value of 0.80 to a final value of 0.90 over 2000 steps, with pruning applied every 500 steps. Continuous experiments were carried out in order to decide these parameters, as a more aggressive pruning process resulted in less accurate models. The pruned models retained their 32-bit floating-point number precision. The CNN model's size decreased by 21.75%, while the CNN+GRU and CNN+LSTM models reduced by 23.81% and 23.19%, respectively.

*4.3. Experimental setup*

We conducted a performance evaluation of the three deep-learning models that we proposed for eye tracking. Our objective is to assess the trade-offs between model accuracy, inference time, and resource usage for baseline, quantised and pruned models. This will help to determine the suitability of each model for specific deployment scenarios, particularly in the context of real edge intelligence applications. Further, we evaluated the RMSE of these optimised models to identify the trade-offs between maintaining model accuracy and achieving optimal efficiency.

To conduct this evaluation, we tested four edge devices: a Raspberry Pi 4 (RPi 4), Odroid N2+, Intel NUC, and an Nvidia Jetson AGX. The key specifications of these devices (CPU, GPU, memory, storage, and operating system) are presented in Table 3. The selection of these devices was based on their diverse hardware capabilities and popularity within the edge computing domain, allowing us to comprehensively evaluate the performance of our eye-tracking models and draw meaningful conclusions regarding their suitability for various deployment scenarios and edge intelligence applications.

In this study, we developed a Python-based program to perform inference on four selected edge devices. This program was designed to measure both the inference time and the time allocated to each step within the processing pipeline. Simultaneously, we created another Python program to assess the memory and CPU usage. This second program operated concurrently with the first, continuously (for each second) monitoring the memory and CPU usage of the primary program by tracking its process ID. For our experimentation, we carefully selected random videos from the dataset created in Section 3.1.

Furthermore, we employed a Rohde and Schwarz NGM202 Two-Quadrant Programmable Power Supply to precisely gauge the power consumption of each model as they operated on various edge devices. Another Python program, running on a separate device, was utilised to interface with the Power Supply, continuously retrieving data at a rate of 10 readings per second. This data acquisition process encompassed simultaneous measurements of voltage (in volts, V) and current (in amperes, A). Subsequently, these collected data were used to compute the energy consumption of each model, expressed in milliwatt-hours (mWh).

**Table 4**

Comparison of processing pipeline and inference timing (in milliseconds per frame) for Baseline, Quantised, and Pruned Models on four Edge Devices.

| Model | Device | Initial Steps | | | Baseline model | | Quantised model | | Pruned model | |
|---|---|---|---|---|---|---|---|---|---|---|
| | | Fr. Read | Face | Preproc. | Infer. | Total | Infer. | Total | Infer. | Total |
| CNN | Odroid N2+ | 4.04 | 86.10 | 1.96 | 181.22 | 273.32 | 56.24 | 148.34 | 162.05 | 254.15 |
| | RPi 4 | 6.69 | 109.21 | 2.66 | 257.91 | 376.47 | 67.84 | 186.4 | 220.14 | 338.7 |
| | Intel NUC | 1.45 | 35.59 | 0.86 | 84.89 | **122.79** | 10.52 | **48.42** | 75.24 | **113.14** |
| | Jetson AGX | 2.42 | 53.90 | 3.19 | 102.82 | 162.33 | 27.51 | 87.02 | 89.91 | 149.42 |
| CNN+GRU | Odroid N2+ | 4.03 | 86.20 | 1.97 | 334.26 | 426.46 | 223.85 | 315.73 | 289.58 | 381.78 |
| | RPi 4 | 6.15 | 108.85 | 2.46 | 433.74 | 551.2 | 255.07 | 369.53 | 370.61 | 488.07 |
| | Intel NUC | 1.66 | 35.42 | 0.97 | 114.38 | **152.43** | 85.62 | **122.70** | 96.33 | **134.38** |
| | Jetson AGX | 2.36 | 53.50 | 3.08 | 120.69 | 179.63 | 113.53 | 182.18 | 116.4 | 175.34 |
| CNN+LSTM | Odroid N2+ | 4.04 | 86.13 | 1.94 | 335.35 | 427.46 | 224.83 | 316.93 | 293.23 | 385.34 |
| | RPi 4 | 6.15 | 109.08 | 2.46 | 442.56 | 560.25 | 257.39 | 371.85 | 385.15 | 502.84 |
| | Intel NUC | 1.51 | 34.52 | 0.82 | 130.48 | **167.33** | 94.05 | **130.99** | 109.40 | **146.25** |
| | Jetson AGX | 3.01 | 56.58 | 3.97 | 136.99 | 200.55 | 124.51 | 183.76 | 128.33 | 191.89 |

*Note:* Fr. Read — Frame Read Time; Face — Face Detection Time; Preproc. — Preprocessing Time; Infer. — Inference Time; Total — Total Time.

## 5. Performance evaluation

This section provides an analysis of the performance of the developed deep learning models for eye tracking on edge devices. The evaluation focuses on key metrics such as inference time, CPU and memory usage, and energy consumption. The trade-offs between model accuracy and optimisation techniques, such as quantisation and pruning, are examined to determine their impact on performance and resource efficiency. Finally, the limitations of the experimental setup and findings are discussed.

### 5.1. Inference time

Inference time is the main performance metric as it directly impacts the model's ability to respond to incoming data, making it an essential factor in determining its suitability for real-time and time-sensitive applications. In this experiment, we measure the time each step takes in the processing pipeline, including frame reading, face detection, preprocessing, and inferencing on four edge devices. This analysis aimed to identify potential processing bottlenecks. Table 4 visualises the inference time and total time taken by baseline, quantised and pruned models for four edge devices used in the previous iteration.

Among the devices tested with the CNN model (baseline), the Intel NUC performed best in all measured categories, showing the lowest time for initial steps (1.45 ms), face reading (35.59 ms), preprocessing (0.86 ms), inference (84.89 ms), and total time (122.79 ms). The Jetson AGX achieved preprocessing and inference times of 3.19 ms and 102.82 ms, respectively, summing up to 162.33 ms. The Odroid N2+ and RPi 4 have their trade-offs between cost and performance when using them for smartphone-based eye tracking, where the Odroid N2+ yielded a total time of 273.32 ms, while the RPi 4 produced 376.47 ms.

For the CNN+GRU model (baseline), the Intel NUC produced the lowest times across all categories, with an inference time of 114.38 ms and a total time of 152.43 ms. The Jetson AGX performed better than the Odroid N2+ and RPi 4, with a total time of 179.63 ms, but was still less efficient than the Intel NUC. Similar to the previous two models, the Intel NUC outperforms the other devices in the baseline CNN+LSTM model. It achieved the lowest times in initial steps, face reading, preprocessing, and inference, culminating in a total time of 167.33 ms. The Jetson AGX takes longer for preprocessing and inference compared to its performance with the CNN model, leading to a total time of 200.55 ms. The Odroid N2+ and RPi 4 continue to lag, with total times of 427.46 and 560.25 ms, respectively, indicating the substantial processing demands of LSTM layers.

The primary reason for this achievement in Intel NUC is due to the Intel Core i7-10710U processor, which is notably more performant than the ARM-based processors found in the other three devices. Further, this performance may be attributed to its overall computational strength and higher power consumption. Additionally, the Jetson AGX appeared to be operating in a lower power-envelope mode, suggesting it might achieve faster performance if not constrained by power limitations. This might indicate the suitability for real-time applications where fast inference is needed; this is made capable through GPU-accelerated processing by Jetson AGX but not equal to Intel NUC.

According to the results of this experiment, for the quantised CNN model, the Odroid N2+ device shows a reduction in inference time from 181.22 ms to 56.24 ms, representing a 69% improvement. Also, the pruned CNN model reduces the inference time to 162.05 ms, a 10.6% improvement. Similarly, on the RPi 4, the quantised CNN model reduces inference time by 73.7%, while the pruned CNN model shows a 14.7% improvement. The Intel NUC and Jetson AGX also exhibit considerable improvements with the quantised CNN model, showing reductions of 87.6% and 73.2%, respectively, while the pruned model improves by 11.4% and 12.6%. Therefore, quantisation consistently offers more significant reductions in inference time compared to pruning across different edge devices for the CNN model.

For the CNN+GRU model, the quantised model on the Odroid N2+ decreases inference time from 334.26 ms to 223.85 ms, whereas the pruned model reduces it to 289.58 ms. On the RPi 4, the quantised model reduced from 433.74 ms to 255.07 ms, and with the pruned model to 370.61 ms. This is a 41.2% improvement with the quantised model and a 14.5% improvement with the pruned model. On the Intel NUC, the quantised model improves inference time by 25.2% and the pruned model by 15.8%. The Jetson AGX shows improvement with the quantised model, reducing from 120.69 ms to 113.53 ms and with the pruned model to 116.4 ms.

For the CNN+LSTM model, Odroid N2+ and RPi 4 exhibit 33% and 41.8% improvement in the inference time. The Intel NUC shows a 27.9% improvement with the quantised model, which is a 2.7% decrease compared to the CNN+GRU model on Intel NUC. On the Jetson AGX, the quantised model improved the inference time from 136.99 ms to 124.51 ms, and the pruned model achieved an inference time of 128.33 ms. The reason for getting similar results for both CNN+GRU and CNN+LSTM might be the similarities in the computational patterns and data dependencies of GRU and LSTM layers.

The performance improvement is comparatively lower for optimised CNN+RNN models than for optimised CNN. Because the recurrent nature of GRU and LSTM layers needs to maintain their state across time steps, this can reduce the efficiency gains through quantisation. In contrast, CNN layers contain more parallelisable operations such as convolution, activation and pooling operations, which can increase the efficiency of the quantised model.

Overall, these results indicate that quantisation consistently achieves significantly greater inference time reductions than pruning. This can be because quantisation simplifies arithmetic operations by reducing data precision, directly speeding up computations. In contrast, pruning reduces the model size by eliminating less important weights and neurons, reducing the number of computations to a lesser extent than quantisation. Both techniques make models more efficient and suitable for deployment on edge devices.

Real-time eye tracking requires a frame rate of at least 30 fps. This means that the total processing time must be under approximately 33 ms per frame. The results show that only the quantised CNN model on the Intel NUC comes close to meeting this requirement. This combination achieved a total time of 48.42 ms, which can perform around 20fps in a real-time smartphone-based eye-tracking application. However, low accuracy in the CNN model may still be a problem for these kinds of applications. The maximum frame rates for the CNN+LSTM and CNN+GRU models are achieved with the quantised version on the Intel NUC, which are 7.63 fps and 8.15 fps, respectively.

Further, it became evident that inference time alone is not the sole obstacle to achieving real-time processing speeds within the pipeline. Among the various preprocessing steps, we identified face detection as a notably time-intensive process. Utilising the Dlib face detection algorithm, we determined that the Intel NUC exhibited the shortest average face detection time, surpassing the RPi 4 (which performed the slowest) by a substantial margin of 67.45%. However, it is essential to acknowledge the trade-off between the accuracy and speed of face detection algorithms. Less precise face detection algorithms may lead to a higher incidence of missing values within the eye-tracking algorithm. One of the future directions is to use fast and tiny face detection algorithms such as YuNet [37] and Haar Classifiers [38]. Moreover, it is important to note that we did not test how the face detection algorithm performs with different cosmetics and face masks, which could impact detection accuracy. Future work could explore different face detection algorithms, such as [39,40].

Disparities in frame reading times may, in part, be attributed to differences in storage speed, RAM bandwidth, and CPU performance across the various edge devices. Additionally, the preprocessing step, which involves frame resizing, normalisation, and conversion to greyscale, consistently took less than 4 ms across all the devices.

## 5.2. CPU and memory usage

CPU usage is mainly applicable when assessing the models' computational efficiency and suitability for deployment on various hardware platforms. High CPU usage can lead to processes being queued, which increases latency and reduces responsiveness. Additionally, high CPU usage itself can cause increased power consumption, which is a crucial consideration for resource-constrained devices like mobile phones and embedded systems. Memory usage, on the other hand, quantifies the volume of working memory necessary to store and execute the eye-tracking models during inference. This measurement is significant in scenarios where memory resources are limited, such as on smartphones or embedded systems. Excessive memory usage can lead to performance degradation due to memory paging and swapping, where the system moves data between RAM and disk storage to manage memory demands. This can slow down applications and potentially cause crashes due to memory exhaustion. Therefore, finding the right balance between model accuracy and memory efficiency is essential for ensuring smooth and reliable operation across various devices and environments.

The results visualised in Table 5 show the CPU and memory usage for the baseline deep learning model across different edge devices, highlighting the variable efficiencies and resource requirements of each platform. These results reflect different capabilities and architectures among the tested devices. For the baseline CNN model, Odroid N2+ and RPi 4 have shown CPU usage of 174.95% and 178.15%, respectively, due to relatively low processing power. Also, this indicates that they require more CPU cycles to handle the computations. The memory usage of the RPi 4 is considerably high at 1751 MB compared to that of the Odroid N2+ at 1356.95 MB. In contrast, the Intel NUC, with a more powerful processor, exhibits a lower CPU usage at 61.66%, which was measured across all cores, indicating that computations are handled efficiently. However, the memory usage on the Intel NUC is high at 1949.62 MB, possibly due to differences in compiler output for the x86 architecture or varying OS process requirements between Windows 11 and Linux.

The CPU and memory usage of the baseline CNN+GRU and baseline CNN+LSTM models reflect the additional computational demands introduced by the recurrent units on top of the convolutional units. Similar to the CNN model, for the CNN+GRU model, the Odroid N2+ and Raspberry Pi 4 exhibit high CPU usage (178.81% and 177.46% respectively) and high memory usage (1880.71 MB and 1547.68 MB, respectively). Regarding the CNN+LSTM model, the Odroid N2+ and Raspberry Pi 4 again show high CPU usage (179.38% and 177.63%, respectively) and high memory usage (1892.9 MB and 1534.16 MB). Moreover, the Intel NUC again outperformed all the edge devices in CPU and memory usage when running both baseline CNN+LSTM and baseline CNN+GRU models. Therefore, the variations in CPU and memory usage across these models and devices highlight the impact of hardware specifications and model complexity on performance.

Table 5
Resource utilisation per frame for baseline, quantised and pruned models on various edge devices.

| Model | Device | Baseline model | | Quantised model | | Pruned model | |
|---|---|---|---|---|---|---|---|
| | | CPU % | Memory (MB) | CPU % | Memory (MB) | CPU % | Memory (MB) |
| CNN | Odroid N2+ | 174.95 | **1356.95** | 106.64 | 417.09 | 144.13 | **1215.65** |
| | RPi 4 | 178.15 | 1751 | 118.47 | 383.61 | 161.61 | 1420.82 |
| | Intel NUC | **61.66** | 1949.62 | **60.69** | 309.33 | **59.66** | 1699.62 |
| | Jetson AGX | 107.58 | 3884.91 | 135.42 | 1124.47 | 105.58 | 3184.91 |
| CNN+GRU | Odroid N2+ | 178.81 | **1547.68** | 103.33 | 440.17 | 146.51 | **1340.63** |
| | RPi 4 | 177.46 | 1880.71 | 109.07 | 403.18 | 166.58 | 1530.73 |
| | Intel NUC | **67.59** | 2129.81 | **64.13** | **327.07** | **61.25** | 1839.52 |
| | Jetson AGX | 124.41 | 4126.78 | 116.95 | 1800.15 | 107.18 | 3376.1 |
| CNN+LSTM | Odroid N2+ | 179.38 | **1534.16** | 103.37 | 440.42 | 147.2 | **1351.8** |
| | RPi 4 | 177.63 | 1892.9 | 108.99 | 403.09 | 167.1 | 1538.38 |
| | Intel NUC | **67.26** | 2135.13 | **64.88** | **328.45** | **65.09** | 1847.7 |
| | Jetson AGX | 110.44 | 4174.27 | 131.55 | 1757.09 | 107.49 | 3398.92 |

With regard to the model complexities, transitioning from the CNN model to the CNN+GRU and CNN+LSTM models (baseline) increases the CPU usage by 2.21% and 2.54%, respectively, on the Odroid N2+ device. Also, memory usage increased by 14.06% and 13.07%, respectively, on the low-end Odroid N2+ device. The CPU usage of the Intel NUC increased by 9.61% and 9.08%, respectively, and memory usage increased by 9.24% and 9.51%, respectively, when transitioning from the CNN model to the CNN+GRU and CNN+LSTM models (baseline). The main reason for this disparity is that the Odroid N2+ already operates near its maximum CPU capacity, leaving less headroom for additional processing demands. In contrast, the NUC has more processing power available to accommodate the increased computational load.

For the CNN model, quantisation leads to a notable reduction in both CPU usage and memory usage. On the Odroid N2+, the CPU usage drops from 174.95% to 106.64%, and memory usage decreases from 1356.95 MB to 417.09 MB. Similar trends are observed on the RPi 4 and Intel NUC, where the quantised model significantly reduces CPU usage and memory usage. The Jetson AGX shows a different pattern with a slight increase in CPU usage but a considerable decrease in memory usage, indicating that while the computational complexity remains high, the memory footprint is significantly reduced.

For the CNN+GRU and CNN+LSTM models, the quantisation reduces both CPU and memory usage across all devices. On the Intel NUC, the quantised CNN+GRU model reduces CPU usage from 67.59% to 64.13% and memory usage from 2129.81 MB to 327.07 MB. The Jetson AGX shows a similar pattern, with memory usage dropping significantly from 4126.78 MB to 1800.15 MB. Moreover, pruned models for CNN+GRU and CNN+LSTM also show improvements, but again, these are less significant compared to quantised models. For example, the pruned CNN+LSTM model on the Odroid N2+ reduces CPU usage from 179.38% to 147.2% and memory usage from 1534.16 MB to 1351.8 MB.

Pruned models have less reduced memory usage compared to quantised models because the remaining weights and neurons of the pruned model still retain their original precision, typically 32-bit floating point. This means that while the model's structure becomes lighter, the memory required to store each weight remains unchanged. Therefore, pruning is not the best optimisation technique for smartphone-based eye-tracking applications concerning memory usage.

Moreover, improving CPU and memory usage is necessary for improving the performance and efficiency of deep learning models on edge devices. Optimising models for these parameters ensures better resource management, leading to faster processing times, reduced latency, and overall improved user experience. Therefore, consideration of hardware specifications and model complexity is essential in the deployment of deep learning models on edge devices, as it directly impacts performance and efficiency.

### 5.3. Energy consumption

Energy consumption measures how efficiently edge devices utilise electrical power to perform their tasks. Minimising energy consumption while maintaining adequate performance is a key challenge in designing and deploying energy-efficient edge intelligence systems. Table 6 provides a comprehensive overview of energy consumption measurements across the four edge devices. The energy consumption metrics are categorised into "total energy" and "additional energy." Total energy signifies the energy the entire system expends to execute inference for a single frame. This metric encompasses all energy utilisation during the inference process, including any other energy consumption by the device (e.g., by background processes, operating system, idle peripherals, etc.). On the other hand, additional energy represents the specific energy consumed exclusively for inference tasks within a single frame. This value is derived by subtracting the energy consumed by the device when idle for an equivalent length of time from the total energy consumption.

The results suggest that the baseline CNN model achieved the lowest total and additional energy consumption per frame on Odroid N2+, which are 0.6038 mWh and 0.1917 mWh, respectively. For the baseline CNN+GRU model, the Jetson AGX exhibits the lowest energy consumption, with a total energy consumption of 0.9333 mWh and an additional energy consumption of 0.2539 mWh.

Table 6
Energy consumption per frame for baseline, quantised and pruned models on various edge devices. The energy consumption is broken down into the total energy (Total) usage and the additional energy (Additional) used specifically for the eye-tracking process (excluding idle energy consumption).

| Model | Device | Baseline model | | Quantised model | | Pruned model | |
|---|---|---|---|---|---|---|---|
| | | Total (mWh) | Additional (mWh) | Total (mWh) | Additional (mWh) | Total (mWh) | Additional (mWh) |
| CNN | Odroid N2+ | 0.6038 | **0.1917** | 0.1898 | **0.0548** | 0.5987 | 0.1770 |
| | RPi 4 | 0.7053 | 0.2954 | 0.2120 | 0.0893 | 0.6894 | 0.2832 |
| | Intel NUC | 1.0727 | 0.8784 | 0.4138 | 0.3326 | 1.0369 | 0.8497 |
| | Jetson AGX | 0.7985 | 0.1970 | 0.2507 | 0.0549 | 0.6941 | **0.1520** |
| CNN+GRU | Odroid N2+ | 1.0173 | 0.3380 | 0.3897 | **0.1095** | 0.9987 | 0.3015 |
| | RPi 4 | 1.1540 | 0.4850 | 0.4105 | 0.1659 | 1.1109 | 0.4649 |
| | Intel NUC | 1.3309 | 1.0904 | 1.1137 | 0.9055 | 1.3322 | 1.0458 |
| | Jetson AGX | 0.9333 | **0.2539** | 0.6926 | 0.2335 | 0.9690 | **0.2486** |
| CNN+LSTM | Odroid N2+ | 1.0280 | **0.3414** | 0.3949 | **0.1104** | 0.9997 | **0.3215** |
| | RPi 4 | 1.1674 | 0.4920 | 0.4172 | 0.1682 | 1.1245 | 0.4781 |
| | Intel NUC | 1.4897 | 1.2285 | 1.1791 | 0.9596 | 1.3888 | 1.1984 |
| | Jetson AGX | 0.9856 | 0.5882 | 0.8203 | 0.3005 | 1.3202 | 0.4247 |

Also, for the baseline CNN+LSTM model, the Jetson AGX has a total energy consumption of 0.9856 mWh and an additional energy consumption of 0.5882 mWh, which is the lowest compared to other devices. This is because the Jetson AGX's GPU capabilities and optimised hardware for deep learning tasks enable it to effectively process the more complex LSTM and GRU units while maintaining lower energy consumption.

The Odroid N2+ and the Raspberry Pi 4, which have lower processing power, present relatively lower total and additional energy consumption. However, the transition from the CNN model to the CNN+GRU and CNN+LSTM models reflects additional energy in computations with recurrent units. Particularly, for the Odroid N2+, the energy consumption increased by approximately 68.47% for CNN+GRU and 70.27% for CNN+LSTM. Similarly, for the Raspberry Pi 4, the increase was about 63.62% for CNN+GRU and 65.54% for CNN+LSTM.

For the quantised CNN model, the total energy consumption drops from 0.6038 mWh to 0.1898 mWh, a 68.58% improvement, and additional energy consumption decreases from 0.1917 mWh to 0.0548 mWh, a 71.41% improvement On the Odroid N2+. Similar trends are observed on the RPi 4 and Intel NUC, where the quantised model significantly reduces both total and additional energy consumption by 69.93% and 62.14%, respectively. The Jetson AGX shows a considerable decrease in total energy consumption from 0.7985 mWh to 0.2507 mWh, a 68.61% improvement, with additional energy usage dropping from 0.1970 mWh to 0.0549 mWh, a 72.13% improvement.

For the CNN+GRU and CNN+LSTM models, quantisation reduces both total and additional energy consumption across all devices. On the Intel NUC, the quantised CNN+GRU model reduces total energy consumption from 1.3309 mWh to 1.1137 mWh, a 16.33% improvement, and additional energy consumption from 1.0904 mWh to 0.9055 mWh, a 16.95% improvement. The Jetson AGX shows a similar pattern, with total energy consumption dropping significantly from 0.9333 mWh to 0.6926 mWh, a 25.81% improvement, and additional energy usage decreasing from 0.2539 mWh to 0.2335 mWh, an 8.04% improvement.

Pruned models for CNN+GRU and CNN+LSTM also show improvements in energy consumption, but these improvements are less significant compared to quantised models. This indicates that while pruning effectively reduces the energy required for computations by eliminating less significant weights and neurons, the benefits are more prominent in memory usage rather than energy savings compared to quantisation. However, these results can be used in future when deploying smartphone-based eye-tracking applications on edge devices.

The Intel NUC exhibits the highest total energy consumption for all the models compared to the ARM-based edge devices. This difference can be attributed to the Intel NUC's hardware characteristics, including its Intel Core i7-10710U processor. While delivering strong computational capabilities, this processor generally consumes more power, particularly at high clock speeds. Additionally, the Intel Core-i7 processor is based on the x86_64 architecture and tends to demand greater power compared to ARM-based architectures like the ARM Cortex-A72 or Cortex-A73. The range of these differences in energy consumption points out that consideration must be given not only to the model complexity but also to the device's capabilities in optimising energy-efficient deployments of deep learning models at the edge.

*5.4. Accuracy trade-off*

One of the limitations of model optimisation is the potential loss of accuracy, especially when aggressive optimisation techniques such as quantisation or pruning are applied. We calculated the RMSE for each model before and after optimisation. Table 7 includes RMSE and $R^2$ between the three models under baseline, quantised and Pruned modes. Additionally, Fig. 5 illustrates the trade-off between RMSE and inference time for baseline, quantised and pruned models on Intel NUC.

The transition from baseline to quantised conditions resulted in an increase in RMSE across all models. Specifically, for the CNN model, the baseline model achieves an RMSE of 1.468 cm and an $R^2$ of 0.83. However, when the model is quantised, the RMSE increases from 1.468 cm to 1.680 cm, and the $R^2$ decreases from 0.83 to 0.78. Similarly, the pruned model shows an RMSE increase to 1.758 cm and an $R^2$ decrease to 0.79. For the CNN+GRU model, quantisation increases the RMSE from 1.091 cm to 1.295 cm, and the $R^2$ decreases from 0.80 to 0.77. Pruning further increases the RMSE to 1.452 cm, with the $R^2$ dropping to 0.78. Similarly, for the CNN+LSTM model, quantisation raises the RMSE from 0.955 cm to 1.192 cm, and the $R^2$ decreases from 0.81 to 0.78. Pruning results in an RMSE of 1.366 cm and an $R^2$ of 0.76. The sequential and interdependent nature of GRU and LSTM layers means that any reduction or alteration in weights through quantisation or pruning can significantly impact accuracy.

**Table 7**
Comparison of RMSE and $R^2$ between Baseline, Quantised and Pruned Models for three Architectures. These models were run on an Intel NUC device.

|  | Baseline model | | Quantised model | | Pruned model | |
| --- | --- | --- | --- | --- | --- | --- |
|  | RMSE (cm) | $R^2$ | RMSE (cm) | $R^2$ | RMSE (cm) | $R^2$ |
| CNN | 1.468 | 0.83 | 1.680 | 0.78 | 1.758 | 0.79 |
| CNN+GRU | 1.091 | 0.80 | 1.295 | 0.77 | 1.452 | 0.78 |
| CNN+LSTM | **0.955** | 0.81 | **1.192** | 0.78 | **1.366** | 0.76 |

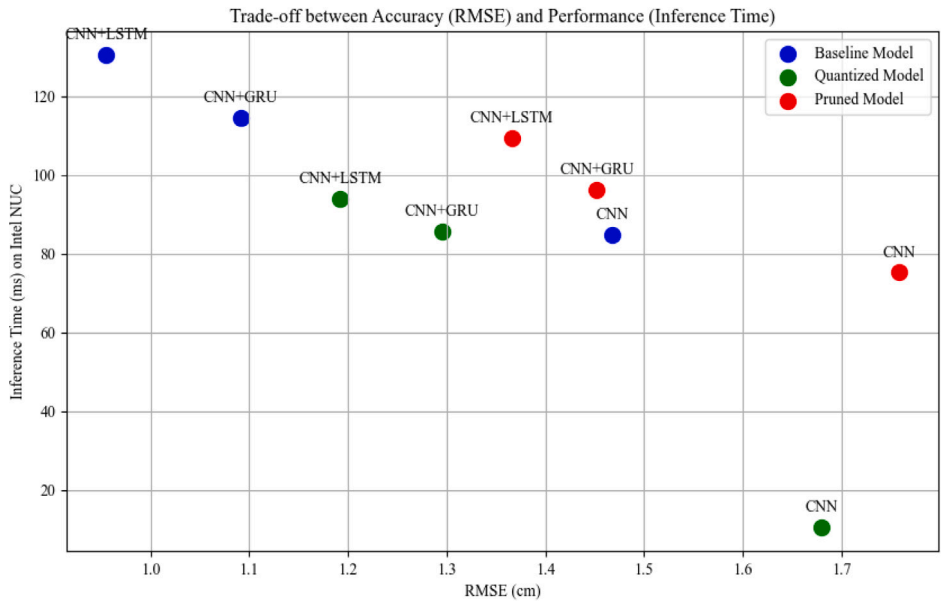

**Fig. 5.** Trade-off between RMSE and inference time for baseline, quantised and pruned models on Intel NUC.

While quantisation and pruning reduce model accuracy, the increases in performance, power consumption and reduced resource requirements may be worth the trade-off in constrained mobile applications. Despite these trade-offs, the significant improvements in inference time, CPU usage, memory usage, and energy efficiency make these optimisation techniques valuable, especially in resource-constrained environments. Therefore, it is essential to emphasise that a balance between accuracy, latency, computational demands, and energy efficiency should guide the selection of an appropriate edge device and model configuration. This approach ensures the development of edge intelligence solutions that are both sustainable and cost-effective.

*5.5. Limitations*

One of the major limitations of this experiment is we could not measure the GPU utilisation of the edge devices with GPUs. Obtaining accurate details of GPU usage is challenging due to several factors. Firstly, many edge devices do not have built-in tools or standardised methods for monitoring GPU power consumption in real-time. While CPUs often have well-documented power management interfaces, GPUs lack such consistent monitoring tools across different manufacturers and models. Secondly, the power consumption of GPUs can fluctuate rapidly based on the workload, making it difficult to capture an accurate and consistent measure without specialised equipment.

Another limitation of this study is the omission of evaluating the communication time between the smartphone and the edge device, which is vital in real-world edge intelligence applications. The communication time can significantly impact the overall system performance and user experience. It can depend on several factors, including the network latency, bandwidth, the efficiency of the communication protocols used, and data compression. Additionally, variations in network conditions, such as interference or congestion, can further influence the communication time. Since these factors introduce more complexity, measurement of the communication time was considered beyond the scope of our controlled experiments.

One limitation in the model optimisation is that we use post-training quantisation instead of quantisation-aware training due to its simplicity. Although this approach is easier to implement and less time-consuming, it may result in greater accuracy loss, as quantisation-aware training can better adapt the model to handle reduced precision during the training process. Additionally, we excluded the evaluation of 8-bit floating point quantised models, considering the high error obtained with 16-bit floating point quantised models.

With regards to the pruned models, we kept their 32-bit floating point precision instead of converting them to a lower precision pruned model. This decision maintained the model's original precision but did not capitalise on potential memory and computational savings that could be achieved through reduced precision. Further improvements could be achieved by combining quantisation and pruning, which would reduce both the model size and the precision of computations.

Further, these models were tested in a controlled environment. Real-world conditions, such as network latency, packet loss, user interaction, and varying input data quality, were not simulated, which could impact the practical applicability of the findings. Finally, the study did not include a comprehensive statistical analysis to determine the significance of the observed differences.

## 6. Conclusion

This study aimed to address the limitations of existing smartphone-based eye-tracking algorithms, particularly their reduced accuracy when applied to dynamic visual stimuli such as video content. We introduced two novel architectures that combine CNN with RNN, specifically LSTM and GRU, to enhance the performance of eye tracking on smartphones. Our models demonstrated improved accuracy, achieving a Root Mean Square Error (RMSE) of 0.955 cm and 1.091 cm for CNN+LSTM and CNN+GRU, respectively, highlighting their effectiveness in handling dynamic stimuli. To overcome the resource constraints inherent in smartphones, we explored edge intelligence as a solution to offload computational tasks from the device to nearby servers. The evaluation of our models on various edge devices, including Raspberry Pi 4, Odroid N2+, Intel NUC, and Nvidia Jetson AGX. Among these devices, Intel NUC was the fastest, processing a single frame in 167.33 ms. Notably, this processing time included 34.52 ms for detecting and extracting the face from a selfie video. These findings helped us find the most suitable edge device for real-time smartphone-based eye tracking and identify the processing pipeline's bottlenecks. We further investigated deep learning model optimisation techniques, such as quantisation and pruning, to enhance energy efficiency and reduce computational overhead on edge devices. Model quantisation reduced the inference time of the CNN+LSTM model from 130.48 ms to 94.05 ms. Also, model pruning reduced the CPU consumption of the Intel NUC by 6.34%. However, quantisation and pruning increased the error rate by 11.03% and 4.69%, respectively. Thus, our results suggest that it is crucial to balance accuracy, latency, computational demands, and energy efficiency when selecting the appropriate edge device and model configuration.

For future work, we recognise the potential for refining our models to achieve higher accuracy and efficiency, particularly when deployed under varying lighting conditions and dynamic user environments. Varying lighting conditions pose challenges due to the impact of illumination on eye visibility and facial feature detection. Different levels of brightness or reflections can affect the model's ability to predict gaze accurately. Future research could involve developing adaptive preprocessing techniques or leveraging advanced data augmentation strategies that simulate diverse lighting scenarios during training. Dynamic user environments introduce variability in factors such as head position, movement, and background changes, all of which can impact gaze estimation accuracy. Future enhancements could include integrating real-time head pose estimation and compensating for motion artifacts to maintain consistent tracking. Exploring hybrid approaches, such as combining quantisation-aware training with dynamic pruning, could balance model efficiency while adapting to these complex scenarios.

## CRediT authorship contribution statement

**Nishan Gunawardena:** Writing – original draft, Software, Methodology, Investigation, Conceptualization. **Gough Yumu Lui:** Writing – review & editing, Supervision, Software, Resources. **Jeewani Anupama Ginige:** Writing – review & editing, Supervision, Project administration. **Bahman Javadi:** Writing – review & editing, Supervision, Resources.

## Declaration of Generative AI and AI-assisted technologies in the writing process

During the preparation of this work the author used ChatGPT 4o in order to improve the readability and language of the manuscript. After using this tool, the authors reviewed and edited the content as needed and take full responsibility for the content of the published article.

## Declaration of competing interest

The authors declare that they have no known competing financial interests or personal relationships that could have appeared to influence the work reported in this paper.

## Data availability

The authors do not have permission to share data.